# Representation of Uncertainty for Limit Processes

**Mark S. Burgin**
Department of Mathematics
University of California, Los Angeles
405 Hilgard Ave.
Los Angeles, CA 90095

## Abstract

Many mathematical models utilize limit processes. Continuous functions and the calculus, differential equations and topology, all are based on limits and continuity. However, when we perform measurements and computations, we can achieve only approximate results. In some cases, this discrepancy between theoretical schemes and practical actions changes drastically outcomes of a research and decision-making resulting in uncertainty of knowledge. In the paper, a mathematical approach to such kind of uncertainty, which emerges in computation and measurement, is suggested on the base of the concept of a fuzzy limit. A mathematical technique is developed for differential models with uncertainty. To take into account the intrinsic uncertainty of a model, it is suggested to use fuzzy derivatives instead of conventional derivatives of functions in this model.

## 1 INTRODUCTION

One of the most popular models in science and engineering is a system of differential equations. Differential equations are used in economics and sociology. These and many other mathematical models utilize limit processes. For example, derivatives in differential equations are constructed as special limits of functions or points. Continuous functions and the calculus, differential equations and topology, all are based on limits and continuity. However, when we perform computations and measurements, we can do only finite number of operations and consequently, achieve only approximate results. At the same time, mathematical technique, e.g., calculus and optimization theory are based on operation of differentiation. This brings us to an unexpected conclusion. Although, it is supposed that numerical computation is a precise methodology in contrast to qualitative methods, this is true only in a very few cases. For limit processes, this is not so and computation adds its uncertainty to the vagueness of initial data. As writes Gregory Chaitin (1999), the fact is that in mathematics , for example, real numbers have infinite precision, but in the computer precision is finite. In some cases, this discrepancy between theoretical schemes and practical actions changes drastically outcomes of a research resulting in uncertainty of knowledge. For example, as remarked the great mathematician Henri Poincare, series convergence is different for mathematicians, who use abstract mathematical procedures, and for astronomers, who utilize numerical computations.

In the paper, a mathematical approach to such kind of uncertainty, which emerges in computation and measurement, is suggested on the base of the concept of a fuzzy limit (Burgin, 1999). A mathematical technique is developed for dealing with differential models with uncertainty. To take into account the intrinsic uncertainty of a model, it is suggested to use fuzzy derivatives instead of conventional derivatives of functions in such models. Two kinds of fuzzy derivatives are introduced: weak and strong ones. Strong fuzzy derivatives are similar to ordinary derivatives of real functions being their fuzzy extensions. At the same time, weak fuzzy derivatives generate a new concept of a weak derivative even in a classical case of exact limits.

Classical results for limits and derivatives, which can be found in textbooks (cf., for example, Goldstein et al,

1987; Shenk, 1979; Ribenboim, 1964; Fihtengoltz, 1955), become direct corollaries of the corresponding results for fuzzy limits and derivatives. These corollaries do not need additional proofs.

The new technique is aimed at better utilization of numerical computations for artificial intelligence, especially in the case when uncertainty of computation is multiplied by the uncertainty of input information.

## 1.1 DENOTATIONS

$\omega$ is the ordered set of all natural numbers; $\varnothing$ is a empty set; **R** is the set of all real numbers; $\mathbf{R}^+ = \{r \in \mathbf{R}; r \geq 0\}$; $\rho(x,y) = |x-y|$ for $x,y \in \mathbf{R}$;

if $l = \{a_i \in M; i \in \omega\}$, and $f: M \to L$, then $f(l) = \{f(a_i); i \in \omega\}$;

if $A = \{a_i; i \in I\}$ is an infinite set, then the expression " a predicate P(x) is true for almost all elements from $A$" means that P(x) can be untrue only for a finite number of elements from $A$. For example, if $A = \omega$, then almost all elements of $A$ are bigger than 10, or another example is that conventional convergence of a sequence $l$ to $x$ means that any neighborhood of $x$ contains almost all elements from $l$.

## 2 MAIN CONCEPTS OF THE THEORY OF FUZZY LIMITS

Here we describe a generalization of the concept of a limit. It is aimed at a reflection of uncertainty that appear in estimation and computation.

Let $r \in \mathbf{R}$ and $l = \{a_i \in \mathbf{R}; i \in \omega\}$.

**Definition 2.1**. A number $a$ is called an $r$-limit of a sequence $l$ (it is denoted $a = r\text{-lim}_{i \to \infty} a_i$ or $a = r\text{-lim } l$) if for any $k \in \mathbf{R}^+ \setminus \{0\}$ the inequality $\rho(a, a_i) \leq r+k$ is valid for almost all $a_i$, i.e., there is such n that for any i>n we have $\rho(a, a_i) \leq r+k$.

**Example 2.1.** Let $l = \{1/i; i \in \omega\}$. Then 1 and -1 are 1-limits of $l$; 1/2 is a (1/2)-limit of $l$ but 1 is not a (1/2)-limit of $l$.

Informally, $a$ is an $r$-limit of a sequence $l$ for an arbitrarily small $k$ the distance between $a$ and all but a finite number of elements from $l$ is smaller than $r+k$.

When $r = 0$, the $r$-limit coincides with the conventional limit of a sequence as the following result demonstrates.
**Lemma 2.1.** $a = \text{lim } l$ if and only if $a = 0\text{-lim } l$.

This result proves that the concept of an $r$-limit is a natural extension of the concept of conventional limit. However, the concept of an $r$-limit actually extends the conventional construction of a limit (cf. Example 2.2).

**Lemma 2.2.** If $a = r\text{-lim } l$ then $a = q\text{-lim } l$ for any $q>r$.

Let $r \in \mathbf{R}$, $l = \{a_i \in \mathbf{R}; i \in \omega\}$, $h = \{b_i \in \mathbf{R}; i \in \omega\}$, $k = \{c_i \in \mathbf{R}; i \in \omega\}$, and $l$ is the disjoint union of $h$ and $k$.

**Lemma 2.3.** $a = r\text{-lim } l$ if and only if $a = r\text{-lim } h$ and $a = r\text{-lim } k$.

**Definition 2.2.** a) A number $a$ is called a fuzzy limit of a sequence $l$ if it is an $r$-limit of $l$ for some $r \in \mathbf{R}$. b) a sequence $l$ fuzzy converges if it has a fuzzy limit. c) the measure of convergence of $l$ to $a$ is equal to $\mu(x=\text{lim } l) = \inf \{r; a \text{ is an } r\text{-limit of } l\}$.

**Example 2.2.** Let us consider three sequences: $l = \{1 + 1/i; i \in \omega\}$, $h = \{1 + (-1)^i; i \in \omega\}$, and $k = \{1 + [(1 - i)/i]^i; i \in \omega\}$. Sequence $l$ has the conventional limit equal to 1 and many fuzzy limits (e.g., 0, 0.5, 2 are 1-limits of $l$). Sequence $h$ does not have the conventional limit but has different fuzzy limits (e.g., 0 is a 1-limit of $h$, while 1, -1, and 1/2 are 2-limits of $h$). Sequence $k$ does not have the conventional limit but has a variety of fuzzy limits (e.g., 1 is a 1-limit of $k$, while 2, 0, 1.5, 1.7, and 0.5 are 2-limits of $k$).

Thus we can see that many sequences that do not have the conventional limit have lots of fuzzy limits.

**Proposition 2.1.** The measure of convergence of $l$ defines the normal fuzzy set Lim $l = [L, \mu(x=\text{lim } l)]$ of fuzzy limits of $l$.

**Theorem 2.1.** If $a = r\text{-lim } l$ and $a > b+r$, then $a_i > b$ for almost all $a_i$ from $l$.

This directly implies the following classical results.

**Corollary 2.1**. If $a = \text{lim } l$ and $a > b$, then $a_i > b$ for almost all $a_i$ from $l$.

**Corollary 2.2**. If $a = \text{lim } l$ and $a > 0$, then $a_i > 0$ for almost all $a_i$ from $l$.

**Corollary 2.3.** If $a_i \leq q$ for almost all $a_i$ from $l$ and $a = r\text{-lim } l$, then $a \leq q + r$.

This directly implies the following classical result.

**Corollary 2.4.** If $a_i \leq q$ for almost all $a_i$ from $l$ and $a = \text{lim } l$, then $a \leq q$.

**Theorem 2.2.** All $r$-limits of a sequence $l$ belong to some interval the length of which is equal to $2r$.

This directly implies the following classical result.

**Corollary 2.5** (any course of mathematical analysis) A limit of a sequence is unique (if this limit exists).

**Theorem 2.3.** A sequence $l$ fuzzy converges if and only if it is bounded.

It gives a criterion for boundedness of a sequence while classical results give only sufficient conditions.

**Remark 2.1.** Not all properties of ordinary limits are properties of fuzzy limits. For example, an *r*-limit may be not unique. In the same way for ordinary (exact) limits, we have the following classical result: if $a_i < a_i$ for almost all $i \in \omega$ and $a = \lim a_i$, $b = \lim b_i$, then $a \leq b$. For fuzzy limits, the resulting inequality is not always true. However, some similar statements for fuzzy limits may be proved.

**Theorem 2.4.** The set $L_r(l) = \{a \in \mathbf{R}; a = r\text{-lim } l\}$ of all *r*-limits of a sequence $l$ is a convex closed set, i.e., $L_r(l) = [a,b]$ for some $a,b \in \mathbf{R}$, or $L_r(l) = \emptyset$ if $l$ has no *r*-limits.

**Theorem 2.5.** If $a = r\text{-lim } l$ and $b = q\text{-lim } h$ then:

a) $a+b = (r+q)\text{-lim}(l+h)$ where $l+h = \{a + b ; i \in \omega\}$;
b) $a-b = (r+q)\text{-lim}(l-h)$ where $l-h = \{a - b ; i \in \omega\}$;
c) $ka = |k| \cdot r\text{-lim }(kl)$ for any $k \in \mathbf{R}$ where $kl = \{ka ; i \in \omega\}$.

This directly implies the following classical result.

**Corollary 2.6** (any course of the calculus). If $a = \lim l$, $b = \lim h$, then:

a) $a+b = \lim (l + h)$;  b) $a-b = \lim (l - h)$;
c) $ka = \lim (kl)$ for any $k \in \mathbf{R}$.

**Corollary 2.7.** If $a$ is a fuzzy limit of $l$ and $b$ is a fuzzy limit of $h$ then $(a+b)$ ( $(a - b)$, and $ka$) is a fuzzy limit of $l + h$ (of $(l - h)$, and $kl$), respectively.

**Definition 2.3.** A sequence $l$ is called *r*-fundamental if for any $k \in \mathbf{R}^+ \setminus \{0\}$ there is such $n \in \omega$ that for any $i, j \geq n$ we have $\rho(a_j, a_i) \leq 2r + k$.

**Definition 2.4.** A sequence $l$ is called fuzzy fundamental if it is *r*-fundamental for some $r \in \mathbf{R}^+$.

**Lemma 2.4.** If $r \leq p$, then any *r*-fundamental sequence is *p*-fundamental.

**Lemma 2.5.** A sequence $l$ is fundamental (in the ordinary sense*)* if and only if it is 0-fundamental.

**Lemma 2.6.** A subsequence of an *r*-fundamental sequence is *r*-fundamental.

**Theorem 2.6 (** *the Extended Cauchy Criterion).* The sequence $l$ has an *r*-limit if and only if it is *r*-fundamental.

**Proof. Necessity.** Let $a = r\text{-lim } l$ and $k \in \mathbf{R}^+ \setminus \{0\}$. Then by the definition $\forall t \in \mathbf{R}^+ \setminus \{0\} \exists n \in \omega \forall i > n \ \rho(a, a_i) \leq r+t$. Consequently, for any $i, j > n$ we have
$\rho(a_i, a_j) \leq \rho(a, a_i) + \rho(a, a_j) \leq 2r + 2t < 2r + k$ if $2t < k$. Thus, $l$ is an *r*-fundamental sequence.

**Sufficiency.** Let $l$ be an *r*-fundamental sequence and $k_m = 1/m$. Then for each number n(m) dependent on m and for all $i, j > n(m)$, we have $\rho(a_i, a_j) \leq 2r + 1/m$, i.e., all points $a_i$ with $i > n(m)$ belong to a closed interval $I_m$ whose length is equal to $2r + 1/m$. Really, let $T_m = \{a_i ; i > n(m)\}$, $b = \sup T_m$, and $c = \inf T_m$. Then all $a_i \in [c, b]$ for all $i > n(m)$.

Let us estimate the length of $I_m$. Suppose that $\rho(b, c) > 2r + 1/m$. It means that $\rho(b, c) \leq 2r + 1/m + h$. At the same time, there are such $a_i, a_j$ for which $i, j > n(m)$, $\rho(b, a_j) \leq (1/10) \cdot h$, and $\rho(c, a_i) \leq (1/10) \cdot h$ because $b$ is the supremum and $c$ is the infimum of all these elements. Consequently, $\rho(a_i, a_j) > 2r + 1/m + (4/5) \cdot h$. It contradicts the choice of the number n(m). Thus, the length of $I_m = [c, b]$ is not bigger than $2r + 1/m$.

We can choose these intervals $I_m$ so that the inclusion $I_{m+1} \subseteq I_m$ will be valid for all m. In such a way, we obtain a sequence of imbedded closed intervals $\{I_m; n \in \omega\}$. The space $\mathbf{R}$ is locally compact. Consequently, the intersection $I = \cap I_m$ is non-void. That is, I consists of one point $d$ or I is a closed interval having the length not bigger than $2r$. When $I = \{d\}$, the sequence $l$ converges, and thus, it is fundamental. This means (by Lemmas 2.4 and 2.5) that $l$ is *r*-fundamental.

When I is a closed interval, its middle $e$ is an *r*-limit of $l$ because for any $k \in \mathbf{R}^+ \setminus \{0\}$ there is some $I_m \supseteq I$ that $k > 1/m$, $\rho(e, e_m) < (1/3) k$ for the center $e_m$ of $I_m$ and almost all $a_i$ belong to $I_m$. Consequently, $\rho(e, a_i) \leq \rho(e_m, a_i) + \rho(e, e_m) < (1/3)k + r + (1/3)k < r + k$, i.e., $e$ is an *r*-limit of $l$.

Theorem is proved.

From Theorem 2.6, we obtain the following result.

**Theorem 2.7 (**the *General Fuzzy Convergence Criterion).* The sequence $l$ fuzzy converges if and only if it is fuzzy fundamental.

This directly implies the following classical result.

**Corollary 2.8 (**the **Cauchy Criterion**). The sequence $l$ converges if and only if it is fundamental.

This result and Lemma 2.5 demonstrate that the concept of an fuzzy convergence is a natural extension of the concept of conventional convergence.

**Proposition 2.2.** The following conditions are equivalent:

1) a sequence $l$ is not fuzzy fundamental;
2) the sequence $l$ is not bounded;
3) some subsequence of $l$ has no fuzzy limits.

## 3  FUZZY DERIVATIVES

Let $X, Y \subseteq \mathbf{R}$, $f: X \to Y$ be a function, $b \in \mathbf{R}$, and $r \in \mathbf{R}^+ \setminus \{0\}$.

**Definition 3.1**. A number $b$ is called a weak centered (left, right, two-sided) *r*-derivative of $f$ at a point $x \in X$ if $b = r\text{-lim } (f(x) - f(x_i)) / (x - x_i)$  (and all $x_i < x$; and all $x_i > x$; $b = r\text{-lim } (f(z_i) - f(x_i)) / (z_i - x_i)$, with $z_i < x < x_i$ for all $i \in \omega$, and the sequences $\{x_i\}, \{z_i\}$ converging to $x$ ) for some sequence $\{x_i\}$ converging to $x$. It is denoted by $b = w_r^{ct} d/_{dx} f(x)$ ($b = w_r^{l} d/_{dx} f(x)$, $b = w_r^{r} d/_{dx} f(x)$, and $b = w_r^{t} d/_{dx} f(x)$, correspondingly ).

**Example 3.1.** Let us take the membership function $m_Q(x)$ of the set of rational numbers, i.e., $m_Q(x)$ is equal to 1 when $x$ is a rational number and $m_Q(x)$ is equal to 0 when $x$ is an irrational number. This function is not even continuous, consequently it does not have derivatives neither in classical sense nor as a generalized function (Shwartz 1950-51). However, at any point $x$ from **R**, $m_Q(x)$ has a weak derivative, which is equal to 0.

**Definition 3.2.** A number $b$ is called a strong centered (left, right, two-sided) $r$-derivative of $f$ at a point $x \in X$ if $b = r\text{-lim} (f(x) - f(x_i))/(x - x_i)$ (and all $x_i < x$; and all $x_i > x$; $b = r\text{-lim} (f(z_i) - f(x_i))/(z_i - x_i)$, with $z_i < x < x_i$ for all $i \in \omega$, and the sequences $\{x_i\}$, $\{z_i\}$ converging to $x$) for all sequences $\{x_i ; i \in \omega\}$ converging to $x$. It is denoted by $b = st_r^{ct} d/_{dx} f(x)$ ($b = st_r^{l} d/_{dx} f(x)$, $b = st_r^{r} d/_{dx} f(x)$, and $b = st_r^{t} d/_{dx} f(x)$, correspondingly ).

**Remark 3.1.** In what follows, $st_r^z d/_{dx} f(x)$ denotes one of these four types of strong and $w_r^z d/_{dx} f(x)$, of weak derivatives of $f(x)$.

**Example 3.2.** Let $f(x) = |x|$. Then $f(x)$ does not have a conventional derivative at 0. However, 0 is the strong two-sided 1-derivative of $f(x)$ at 0, 1 is a strong right, while -1 is a strong left 0-derivatives of $f(x)$ at 0.

**Example 3.3.** Piecewise linear transformations on the interval have been widely studied in the theory of dynamical systems (Collet and. Eckmann 1980), and under different names as well: broken linear transformations (Gervois and Mehta 1977), or weak unimodal maps (Marcuard and Visinescu 1992). An example of such functions if given by a skew tent map $f_{a,b}(x)$ that is equal to $b + ((1-b)/a) \cdot x$ when $0 \leq x < a$ and equal to $(1-x)/(1-a)$ when $a \leq x \leq 1$. Piecewise linear transformations do not have conventional derivatives at some points but they have strong fuzzy derivatives at all points. It provides for application of differential methods to these mappings as well as to dynamics generated by them.

**Remark 3.2.** In contrast to the classical derivative, it is possible that different numbers are strong centered (or left, right, two-sided) $r$-derivative of $f$ at a point $x$.

Let $x$ be an isolated point of $X$.

**Lemma 3.1.** Any number $b \in \mathbf{R}^+$ is a strong centered (left, right, two-sided) $r$-derivative of $f$ at a point $x \in X$ for any $r \in \mathbf{R}$.

**Lemma 3.2.** Any strong centered (left, right, two-sided) $r$-derivative of $f$ at a point $x \in X$ is a weak one-sided (left, right, two-sided) $r$-derivative of $f$ at the same point for any $r \in \mathbf{R}$.

**Lemma 3.3.** $b = st_r^z d/_{dx} f(x)$ if and only if $b = r\text{-lim } E$ where $E = \{\{ (f(x) - f(x_i))/(x - x_i); i \in \omega\}; \{x_i; i \in \omega\}$ is a corresponding sequence converging to $x \}$.

**Lemma 3.4.** Any strong (weak) centered $r$-derivative of $f$ at a point $x \in X$ is both (either) a strong (weak) left and strong (weak) right $r$-derivative of $f$ at the same point for any $r \in \mathbf{R}$.

**Lemma 3.5.** If $b$ is both a strong (weak) left and strong (weak) right $r$-derivative of $f$ at a point $x \in X$, then strong (weak) centered $r$-derivative of $f$ at the same point for any $r \in \mathbf{R}$.

**Proof.** Let us consider a sequence $\{x_i \in \mathbf{R}; i \in \omega\}$ converging to $x \in X$ and let $b$ be both strong (weak) left and strong (weak) right $r$-derivatives of $f$ at $x$. Then the sequence $\{x_i \in \mathbf{R}; i \in \omega\}$ consists of two subsequences $\{v_i \in \mathbf{R}; i \in \omega\}$ and $\{z_i \in \mathbf{R}; i \in \omega\}$ such that $v_i < x$ and $z_i > x$ for all $i \in \omega$. Each of them is either finite or it converges to $x$. When one of these subsequences is finite, then $b = r\text{-lim } (f(x) - f(x_i))/(x - x_i))$.

Let both subsequences $\{v_i \in \mathbf{R}; i \in \omega\}$ and $\{z_i \in \mathbf{R}; i \in \omega\}$ be infinite. By the definition of strong $r$-derivatives $b = r\text{-lim}( f(x) - f(v_i))/(x - v_i))$ and $b = r\text{-lim } (f(x) - f(x_i))/(x - x_i)$. Then by Lemma 2.3, we have $b = r\text{-lim } (f(x) - f(x_i))/(x - x_i))$. As the sequence $\{x_i \in \mathbf{R}; i \in \omega\}$ is chosen arbitrarily, Lemma 3.5 is proved.

**Definition 3.3.** A number $b$ is called a complete $r$-derivative of $f$ at a point $x \in X$ if $b$ is at the same time a strong centered, left, right, and two-sided $r$-derivative of $f$ at the point $x$.

**Proposition 3.1.** If $b$ is a strong centered $r$-derivative of $f$ at a point $x \in X$, then $b$ is a strong two-sided $r$-derivative of $f$ at the point $x \in X$.

**Proof.** Let us consider an arbitrary sequence $\{ (f(z_i) - f(x_i))/( z_i - x_i); z_i > x > x_i, i \in \omega \}$. Geometrical considerations demonstrate that either $(f(x) - f(x_i))/(x - x_i)) \leq ((f(z_i) - f(x_i))/( z_i - x_i)) \leq (f(x) - f(z_i))/(x - z_i)$ or $(f(x) - f(x_i))/(x - x_i)) \geq ((f(z_i) - f(x_i))/( z_i - x_i)) \geq (f(x) - f(z_i))/(x - z_i)$. Consequently, if $b$ is a strong centered $r$-derivative of $f$ at $x$, then $b$ is an $r$-limit of the sequences $\{ (f(x) - f(z_i))/(x - z_i) \}$ and $\{ (f(x) - f(x_i))/(x - x_i) \}$. Properties of $r$-limits imply that $b$ is an $r$-limit of the sequence $\{ ((f(z_i) - f(x_i))/( z_i - x_i)) \}$. As it is considered an arbitrary system, then $b$ is (by the definition) a strong two-sided $r$-derivative of $f$ at the point $x \in X$.

Proposition is proved.

Let $f$ be a continuous function at a point $x \in X$.

**Proposition 3.2.** If a strong two-sided $r$-derivative of $f$ at a point $x \in X$ exists (and is equal to $b$) then both one-sided strong $r$-derivatives of $f$ at a point $x \in X$ exist (and coincide with $b$).

**Proof.** Let us consider a sequence $\{x_i \in \mathbf{R}; i \in \omega\}$ converging to $x \in X$ and all $x_i < x$. As $f$ is a continuous function at $x$, it is possible to correspond to each $x_i$ such $z_i$ that $x < z_i$ and $|x - z_i| < 1/i$. Then $|b - ((f(x) - f(x_i))/(x - x_i))| < |b - ((f(x) - f(x_i))/(z_i - x_i))| + \varepsilon_i < |b - ((f(z_i) - f(x_i))/(z_i - x_i))| + \varepsilon_i$. Both sequences $\{|b - ((f(z_i) - f(x_i))/(z_i - x_i))|\}$ and $\{\varepsilon_i\}$ converge to zero. Consequently, number $b \in \mathbf{R}^+$ is a strong left $r$-derivative of $f$ at the point $x$.

In a similar way, we prove that $b$ is a strong right $r$-derivative of $f$ at the point $x$.

**Corollary 3.1.** If the strong two-sided $r$-derivative of $f$ at a point $x \in X$ exists (and is equal to $b$) then a strong centered $r$-derivative of $f$ at a point $x \in X$ exists (and coincides with $b$).

**Remark 3.3.** Continuity of $f$ is essential for the validity of Proposition 3.2. It is demonstrated by the following example.

Let $f(x) = x$ for all $x > 0$, $f(x) = -x$ for all $x < 0$, and $f(0) = 1$. Then $f$ has a strong two-sided 3-derivative at 0 having no strong one-sided $r$-derivatives at 0 for any $r \in \mathbf{R}$.

Proposition 3.2 imply the following result.

**Corollary 3.2.** If $b$ is a strong centered $r$-derivative of $f$ at a point $x \in X$, then $2b$ is a strong centered $(b+r)$-derivative of $f$ at a point $x \in X$.

From Lemmas 3.4, 3.5, Proposition 3.2, and Corollary 3.1, we obtain the following result.

**Corollary 3.3.** If $b$ is a strong centered $r$-derivative of $f$ at a point $x \in X$, then $2b$ is a strong complete $(b+r)$-derivative of $f$ at a point $x \in X$.

**Corollary 3.4.** If $b$ is a strong centered $r$-derivative of $f$ at a point $x \in X$, then $b$ is a strong complete $(b+2r)$-derivative of $f$ at a point $x \in X$.

**Proposition 3.3.** a) The strong centered 0-derivative of $f$ at a point $x \in X$ is unique and equal to the classical derivative $f'(x)$ of $f$ at $x$. b) The classical derivative $f'(x)$ of $f$ at $x$ is equal to the strong centered 0-derivative of $f$ at $x$.

Proof follows from definitions and uniqueness of $f'(x)$.

This result demonstrates that the concept of a fuzzy derivative is a natural extension of the concept of the conventional derivative.

**Lemma 3.6.** If $b = w_r d^z/_{dx} f(x)$ ($b = st_r d^z/_{dx\_} f(x)$), then $b = w_q d^z/_{dx} f(x)$ ($b = st_q d^z/_{dx} f(x)$) for any $q > r$.

**Proposition 3.4.** If $b$ is a weak (strong) $r$-derivative of $f$ at $x$ and $\rho(b, e) < k$ then $e$ is a weak (strong) $(r+k)$-derivative of $f$ at $x$.

**Corollary 3.5.** If $b = f'(x)$ and $\rho(b, e) < k$ then $e$ is a strong $k$-derivative of $f$ at $x$.

**Proposition 3.5.** If $b$ is a strong $r$-derivative of $f$ at $x$ and is not a strong $k$-derivative of $f$ for any $k < r$, then: a) for any weak $p$-derivative $u$ of $f$ at $x$ the inequality $\rho(b, u) < r+p$ is valid; b) there is exactly one weak 0-derivative $w$ of $f$ at $x$ for which $\rho(b, w) = d$.

**Definition 3.4.** Any $r$-derivative of $f$ at a point $x \in X$ is called a fuzzy derivative of $f$ at the same point and of the same type (i.e., weak, strong, centered, right, left, or two-sided).

It is denoted $b = wd^z/_{dx} f(x)$ ($b = std^z/_{dx} f(x)$).

From Proposition 3.4., we have the following result.

**Corollary 3.6.** If $b = f'(x)$, then $b = std^{ct}/_{dx} f(x)$.

Let $x$ be a non-isolated point of $X$.

**Corollary 3.7.** If $b = std^z/_{dx} f(x)$, then $b = wd^z/_{dx} f(x)$.

Let $WCFD_r(f,x)$ ($WLFD_r(f,x)$, $WRFD_r(f,x)$, $WTFD_r(f,x)$, $SCFD_r(f,x)$, $SLFD_r(f,x)$, $SRFD_r(f,x)$, $STFD_r(f,x)$) be the set of all weak centered (weak left, weak right, weak two-sided, strong centered, strong left, strong right, strong two-sided) $r$-derivatives of $f$ at a point $x \in X$. In what follows, $YXFD_r(f,x)$ denotes one of these sets (i.e., Y may be equal to W or S, while X may be equal to C, L, R, T) and is called the complete $r$-derivative of $f$ at a point $x \in X$ having type Y,X.

**Example 3.4.** Let $f(x) = |x|$. Then $SCFD_0(f,1) = \{1\}$, $SCFD_0(f,0) = [-1, 1]$, and $SCFD_1(f,1) = [0,2]$.

**Theorem 3.1.** Each set $SXFD_r(f,x)$ is a convex closed set, i.e., $SXFD_r(f,x) = [a,b]$ for some $a, b \in \mathbf{R}$, or $SXFD_r(f,x) = \emptyset$ if $f$ has no strong $r$-derivatives of the type X.

**Theorem 3.2.** The following conditions are equivalent:

1) a function $f$ has a strong fuzzy derivative at $x$ of the type X;

2) the sets $WXFD_r(f,x)$ are non-empty and bounded for all $r \geq 0$;

3) there is such $t \geq 0$ that the sets $SXFD_r(f,x)$ are non-empty for all $r \geq t$;

4) the set $WXFD_0(f,x)$ is non-empty and bounded.

**Proposition 3.6.** The set $WCFD_0(f,x)$ consists of a single point if and only if the classical derivative $f'(x)$ exists.

Lemma 3.2 implies the following result.

**Corollary 3.8.** $SXFD_r(f,x) \subseteq WXFD_r(f,x)$ when X is equal to C, L, R, T.

Theorems 2.2 and 2.3 imply the following results.

**Proposition 3.7.** The set $YXFD_r(f,x)$ is a join of closed intervals.

**Proposition 3.8.** If $r \leq p$, then $YXFD_r(f,x) \subseteq YXFD_p(f,x)$.

**Proposition 3.9.** If $b$ is a weak (strong) $r$-derivative of $f$ at $x$, then $\rho(b, WXFD_0(f,x)) \leq r$.

**Corollary 3.9.** If $b$ is a weak (strong) $r$-derivative of $f$ at $x$, then $\rho(b, WXFD_k(f,x)) \leq r-k$ where $r-k = r-k$ when $r \leq k$, otherwise $r-k=0$.

The sets $YXFD_r(f,x)$ define complete global $r$-derivatives $YXFD_r f$ of $f$ on $\mathbf{R}$. Each $YXFD_r f$ is a binary relation on $\mathbf{R}$, and namely, $YXFD_r f = \{ (x,z); x \in \mathbf{R}, z \in YXFD_r(f,x) \}$.

**Proposition 3.10.** The sets $YXFD_r f$ are closed for all $r \geq 0$.

From Theorem 2.5, we obtain the following result demonstrating local linearity and additivity of strong fuzzy derivatives.

**Theorem 3.3.** *a)* If $b$ is a strong centered (left, right, two-sided) $a$-derivative of $f$ at $x$ and $c$ is a strong $d$-derivative of $g$ at $x$, then $b \pm c$ is a strong centered (left, right, two-sided) $(a+d)$-derivative of $f \pm g$ at $x$.

b*)* If $b$ is a strong centered (left, right, two-sided) $a$-derivative of $f$ at $x$ and $r \in \mathbf{R}$, then $r \cdot b$ is a strong centered (left, right, two-sided) $|r| \cdot a$-derivative of $r \cdot f$ at $x$.

**Remark 3.4.** When the conditions of Theorem 3.3 are satisfied, the point $b - c$ is not necessarily a strong $(a - d)$-derivative of $f - g$ at $x$. However, in some cases, it might be such a derivative.

As a consequence, Theorem 3.3 gives the well-known result of the classical analysis.

**Corollary 3.10.** a) If $b = f'(x)$ and $c = g'(x)$, then $b \pm c = (f \pm g)'(x)$. b) If $b = f'(x)$ and $r \in \mathbf{R}$, then $rb = rf'(x)$.

**Corollary 3.11.** The set $SXFD(f,x) = \bigcup_{r \geq 0} SXFD_r(f,x)$ of all fuzzy derivatives of $f$ at $x$ is a real linear space.

**Corollary 3.12.** a) (Global additivity) If $b = \text{std}_r{}^z/_{dx} f$ and $c = \text{std}_q{}^z/_{dx} g$, then $b \pm c = \text{std}_{r+q}{}^z/_{dx} (f \pm g)$.

b) (Global uniformity) If $b = \text{std}_r{}^z/_{dx} f$ and $a \in \mathbf{R}$, then $ab = \text{std}_r{}^z/_{dx} af(x)$.

**Corollary 3.13.** The set $SXFD(f) = \bigcup_{r \geq 0} SXFD_r(f)$ of all fuzzy derivatives of $f$ is a real linear space.

**Remark 3.5.** For weak fuzzy derivatives, Theorem 3.3.a is invalid as the following example demonstrates.

Let $f(x)=1$ when $x \neq u_n=1/n$; $f(x)=1/n$ when $x=u_n$, and $g(x)=1$ when $x \neq v_n=1/2n$; $g(x)=1/n$ when $x=v_n$.

Then 1 is a weak 0-derivative of $f$ and g at 0, but 1+1=2 is not a weak (0+0)-derivative of $f+g$ at 0.

However, for weak fuzzy derivatives, it is possible to deduce some weaker properties of additivity than those possessed by the strong fuzzy derivatives.

Let us assume that: 1) $f: X \to \mathbf{R}$ and $g: X \to \mathbf{R}$ are arbitrary real functions; 2) the sets $\{wd_a{}^z/_{dx} f(x)\}$ of all weak $a$-derivatives of $f$ at $x$ and $\{wd_d{}^z/_{dx} g(x)\}$ of all weak $d$-derivatives of $g$ at $x$ are bounded; and 3) $\sup\{wd_a{}^z/_{dx} f(x)\} = u$, $\sup\{wd_d{}^z/_{dx} g(x)\} = v$.

**Proposition 3.11.** If $b$ is a weak $a$-derivative of $f$ at $x$ and $c$ is a weak $d$-derivative of $g$ at $x$, then there is such $e \in \mathbf{R}$ that $e$ is a weak $(a+d)$-derivative of $f+g$ at $x$ and $e \leq \min\{b+v; c+u\}$.

# 4 FUZZY DIFFERENTIABLE FUNCTIONS

**Definition 4.1.** A function $f$ is fuzzy differentiable (from the left, from the right) at a point $x$ from $X$ if there is some number $a$ such that $f$ has a strong centered (strong left, strong right) $a$-derivative at $x$.

**Remark 4.1.** There are such functions that have no derivative at any point of R but are fuzzy differentiable at all points of R. To demonstrate this, let us consider the function $f(x)$ defined by the following formula:

$f(x) = \Sigma_{n-1}{}^\infty g(4^{n-1}x)/4^{n-1}$ where $g(x + n) = |x|$ for all $x$ with $|x| \leq 1/2$.

It is demonstrated in (Gelbaum and Olmsted 1964) that this function has no derivative at any point of R. At the same time, it is possible to prove that 0 is a strong centered and two-sided 5-derivative of f at any point $x$ from $\mathbf{R}$.

Theorem 3.2 provides for the following criterion of fuzzy differentiability.

**Proposition 4.1.** A function $f$ is fuzzy differentiable (from the left, from the right) at a point $x$ from $X$ if and only if) the set $WXFD_0(f,x)$ is non-empty and bounded.

**Theorem 4.1.** If a function $f$ is fuzzy differentiable at a point $x$ from $X$, then there is such a minimal number $a$ that $f$ has a strong centered $a$-derivative at $x$, i.e., $a = \min\{r; SCFD_r(f,x) \neq \varnothing\}$.

**Proof.** Let us consider the set $FD(f,x) = \{r; f$ has a strong centered $r$-derivative at $x\}$ and the number $a = \inf FD(f,x)$. If $c$ is a number from $FD(f,x)$, then for any sequence $l = \{(f(z_i) - f(x_i))/(z_i - x_i); z_i > x > x_i, i \in \omega, x = \lim x_i = \lim z_i\}$ there is such a point $u$ in $X$ that

$u = c$-lim $l$. Set FD $l = \{ r; l$ has an $r$-fuzzy limit$\}$ is a closed ray. By the definition of a strong centered fuzzy derivative, FD($f,x$) = $\cap$ FD $l_t$ for all such sequences $l_t$ that have the form of the initial sequence $l$. Any intersection of closed rays is a closed set. It may be void, but in our case FD($f,x$) $\neq \emptyset$. Consequently, it is a closed ray of positive numbers. As a closed set, FD($f,x$) contains $a$, q.e.d.

**Corollary 4.1.** A function $f$ has the classical derivative at $x$ if and only if FD($f,x$) = 0.

**Corollary 4.2.** A function $f$ is differentiable on $X$ if and only if FD($f,x$) = 0 for all points $x$ from $X$.

These results and some others (e.g., Theorem 4.3) demonstrate that the concept of a fuzzy differentiability is a natural extension of the concept of the conventional differentiability.

**Remark 4.2.** For weak fuzzy derivatives, Theorem 4.1 is invalid.

The sets YXFD$_r$($f,x$), introduced in the previous section, define the complete fuzzy derivative YXFD($f,x$) of $f$ at a point $x \in X$ having type Y,X. It is called also a complete local fuzzy derivative of $f$. Each YXFD($f,x$) is a fuzzy subset of R, and namely, YXFD($f,x$) =(R, $\mu_x$, [0,1]) where the membership function $\mu_x$ is defined by the equality $\mu_x(z) = 1/(1 + m(x,z))$ where $m(x,z) = \min \{r \in R; z \in$ YXFD$_r$($f,x$)$\}$. This minimum exists by Theorem 4.1.

If we take the union of all complete fuzzy derivatives YXFD($f,x$), we obtain the complete global fuzzy derivative YXFD $f$ of $f$ on $X$ having type Y,X. Here YXFD($f,x$) is a fuzzy binary relation on R, and namely, YXFD $f = (R^2, \mu, [0,1])$ where the membership function $\mu$ is defined by the equality $\mu(x,z) = 1/(1 + m(x,z))$. The result of Theorem 4.1 provides for correctness of the definition of the fuzzy sets YXFD($f,x$) for all $x \in R$ as well as of the fuzzy set YXFD $f$. By Corollary 4.1, $\mu_x(z) = \mu(x,z) = 1$ if and only if $z = f'(x)$ at the point $x$. Consequently sets YXFD$_r$($f,x$) and YXFD $f$ are fuzzy set derivatives of crisp (ordinary) functions related to similar constructions from the classical calculus.

Here we can see in an explicit form how investigation of ordinary functions involves construction of fuzzy sets and relations.

**Remark 4.3.** Complete fuzzy derivatives do not possess many properties of ordinary derivatives as well as of other (strong centered, left, right, two-sided etc.) fuzzy derivatives. For example, let us take $f(x) = |x|$ and $g(x) = -|x|$. Then $f + g$ is the function identically equal to zero. All its derivatives are also equal to zero at all points. Consequently, $\mu_0(0) = 1$ for $f + g$.

At the same time, the value of the membership function $\mu_0(0)$ for the sum of any pair of fuzzy sets YXFD($f,0$) and YXFD($g,0$) is equal to 1/2. Thus the complete fuzzy derivative of the sum $f + g$ is not equal to the sum of the complete fuzzy derivatives of $f$ and of $g$. However, it is well-known that the conventional differentiation is a linear operator (J. Dieudonné, 1960) and the same is true for all kinds of (strong centered, left, right, two-sided etc.) strong fuzzy derivatives (cf. Corollary 3.12).

Let us investigate interrelations between different types of fuzzy differentiation and continuity of functions.

**Theorem 4.2.** The following conditions are equivalent:

1) $f$ is a fuzzy differentiable function at a point $x$ from $X$;
2) $f$ has a strong centered fuzzy derivative at $x$;
3) $f$ has strong right and left fuzzy derivatives at $x$;
4) $f$ is continuous at $x$ and has a strong two-sided fuzzy derivative at $x$.

**Proof.** 2) $\Rightarrow$ 4) Let $f$ be a fuzzy differentiable function at a point $x$ from $X$. Then by the definition 4.1, then there is some number $a$ such that $f$ has a strong centered $a$-derivative $b$ at $x$. Proposition 3.1 implies that $b$ is a strong two-sided $a$-derivative of $f$ at $x$. By the definition it means that for any sequence $\{x_n; x_n \in X\}$, if $x = \lim x_n$, then $b = a$-lim $l$ where $l = \{ (f(x) - f(x_n))/(x - x_n); i \in \omega \}$.

By the definition of fuzzy limits, there is some $m \in \omega$ that for all $n > m$ the following inequality is valid:

$\rho( b, (f(x) - f(x_n))/(x - x_n)) \leq a$. It implies the inequality $|(f(x) - f(x_n))/(x - x_n)| \leq a + |b|$. Consequently we have $| f(x) - f(x_n) | \leq (a + |b|)| x - x_n|$. When $x_n \to x$, it implies that $f(x_n) \to f(x)$. It means that the function $f$ is continuous at $x$.

The implication 4) $\Rightarrow$ 3) follows from Proposition 3.2.

The implication 4) $\Rightarrow$ 2) follows from the Corollary 3.1.

All statements 1), 2), and 3) are equivalent by definitions and results of the previous section.

Theorem is proved.

**Corollary 4.3.** If $f$ is a fuzzy differentiable function at a point $x \in X$, then $f$ is continuous at $x$.

This directly implies the following classical result.

**Corollary 4.4**. If $f$ is a differentiable function at a point $x \in X$, then $f$ is continuous at $x$.

As a consequence, we obtain the following result.

**Theorem 4.3.** Any fuzzy differentiable function $f$ is continuous.

This directly implies the following classical result.

**Corollary 4.5**. Any differentiable function *f* is continuous.

## 5 CONCLUSION

Thus, we have demonstrated that in a broader context of fuzzy limits and derivatives, it is possible to extend the majority of basic results of the classical mathematical analysis. This approach is called the neoclassical analysis (Burgin, 1995). Such a transition to a fuzzy context provides for completion of some basic results of the classical mathematical analysis. As an example, we can take Theorem 2.3. This does not only bring new mathematical results, but also produces deeper insights and a better understanding of the classical theory. In addition to this, it makes possible to eliminate discrepancies existing in numerical analysis. The problem is that computations are realized on finite machines, while many processes of mathematics, such as differentiation and integration, demand the use of a limit, which is an infinite process. As a consequence, correct algorithms based on classical methods of calculus, when implemented, turn into unreliable programs. Neoclassical analysis treats such processes more adequately. It is demonstrated by applications of neoclassical analysis to problems of numerical computations and control (Burgin, 1997; Burgin and Westman, 2000). Consequently, the new technique provides for a better utilization of numerical computations for artificial intelligence, especially in the case when uncertainty of computation is multiplied by the uncertainty of input information.

It is necessary to remark that now computer scientists strive to incorporate limit processes into a computational scheme (cf., for example, Moore 1996; Freedman 1998; Burgin,1992; 2000). One of the goals for these efforts is to put classical mathematics on computer. However, it is impossible to do without adjusting classical mathematics to computer media. The development of the neoclassical analysis is a step in this direction.

**References**

M. Burgin (1992) Universal limit Turing machines, *Notices of the Russian Academy of Sciences*, v.325, No. 4: 654-658 (translated from Russian)

M. Burgin, (1995) Neoclassical Analysis: Fuzzy continuity and Convergence, *Fuzzy Sets and Systems*, v. 75: 291-299

M. Burgin (1997) Extended Fixed Point Theorem, *Methodological and Theoretical Problems of Mathematics and Information Sciences*, 3: 71-81 (in Russian)

M. Burgin (2000) Theory of Fuzzy Limits, *Fuzzy Sets and Systems*, 75: 291-299

M. Burgin (2000) Topological Computations, in "*Association for Symbolic Logic, 2000 Annual Meeting*," University of Illinois at Urbana-Champaign, 23

M. Burgin and J. Westman (2000) Fuzzy Calculus Approach to Computer Simulation, in "*Proceedings of the Business and Industry Simulation Symposium*," Washington, 41-46

G.J. Chaitin (1999) *The Unknowable*, Springer-Verlag, Berlin/Heidelberg/New York

P. Collet and J.-P. Eckmann (1980) *Iterated maps on intervals as dynamical systems,* Boston: Birkhauser

J. Dieudonné (1960) *Foundations of Modern Analysis*, New York and London: Academic Press

B.R.Gelbaum, J.M.H. Olmsted, (1964) *Counterexamples in Analysis*, San Francisco, Holden-Day, Inc.

A. Gervois and M.L. Mehta, (1977) Broken linear transformations, *J. Math. Phys.*, v. 18, pp. 1476-1479

L.J.Goldstein, D.C. Lay, and D.I. Schneider (1987) *Calculus and its Applications,* New Jersey,

G.M.Fihtengoltz, (1955) *Elements of Mathematical Analysis*, Moscow, GITTL (in Russian)

M.H. Freedman (1998) Limit, Logic, and Computation, *Proc. Nat Acad. USA*, 95: 95-97

J.C. Marcuard and E. Visinescu, (1992) Monotonicity properties of some skew tent maps, *Ann. Inst. Henri Poincare,* 28: 1-29

M. Misiurewicz (1989) Jumps of entropy in one dimension, *Fund. Math.,* 132: 215-226

C. Moore (1996) Recursion theory on the reals and continuous-time computation: Real numbers and computers," *Theoretical Computer Science,* 162: 23-44

P.Ribenboim, (1964) *Functions, Limits, and Continuity*, New York / London/ Sydney

J.F.Randolph, (1968) *Basic Real and Abstract Analysis,* New York / London

A.Shenk, (1979) *Calculus and Analytic Geometry*, Santa Monica

L.Shwartz, (1950-51) *Theorie des Distributions,* v. I-II, Hermann, Paris